\documentclass{article}
\usepackage{spconf}
\usepackage{amsmath,graphicx}
\usepackage{siunitx}
\usepackage{url}
\usepackage{makecell}
\usepackage{bbm}
\usepackage[labelformat=simple]{subcaption}

\title{Temporally precise action spotting in soccer videos using dense detection anchors}
\twoauthors
  {João V. B. Soares \qquad Avijit Shah}
	{Yahoo Research}
  {Topojoy Biswas\sthanks{Topojoy Biswas performed the work
	while at Yahoo Research.}}
	{Walmart Labs}
\begin{document}
%
\maketitle
\begin{abstract}
We present a model for temporally precise action spotting in videos,
which uses a dense set of detection anchors, predicting a detection confidence and corresponding fine-grained temporal displacement for each anchor. We experiment with two trunk architectures, both of which are able to incorporate large temporal contexts while preserving the smaller-scale features required for precise localization: a one-dimensional version of a u-net, and a Transformer encoder (TE). 
We also suggest best practices for training models of this kind, by applying Sharpness-Aware Minimization (SAM) and mixup data augmentation. We achieve a new state-of-the-art on SoccerNet-v2, the largest soccer video dataset of its kind, with marked improvements in temporal localization. Additionally, our ablations show: the importance of predicting the temporal displacements; the trade-offs between the u-net and TE trunks; and the benefits of training with SAM and mixup.
\end{abstract}
\section{Introduction}
\label{sec:introduction}

The {\it action spotting} task as proposed by Giancola et al.~\cite{giancola2018soccernet}
aims to detect a single characteristic time instant for each action in a video. 
In the current paper, we tackle action spotting on SoccerNet-v2,
which is currently the largest soccer dataset of its kind with respect to several important metrics~\cite{deliege2021soccernet}.

A significant shortcoming of previous soccer action spotting approaches~\cite{tomei2021rms,cioppa2020context,cioppa2021camera,giancola2021temporally,zhou2021feature,vanderplaetse2020improved,vats2020event} is their imprecise temporal localization. 
While temporal localization errors are acceptable when finding only the main events within soccer matches (which tend to have longer durations), there are a variety of applications where they are not. Examples include detecting scoring events in faster-paced sports such as basketball and volleyball, as well as short events within soccer itself, such as {\it ball out of play}, and {\it throw-in} (the most frequent actions in SoccerNet-v2), or {\it passes} and {\it challenges}, which are not in SoccerNet-v2, but are highly relevant for sports analytics.

\begin{figure*}
\begin{subfigure}{.7\linewidth}
  \includegraphics[width=12.0cm]{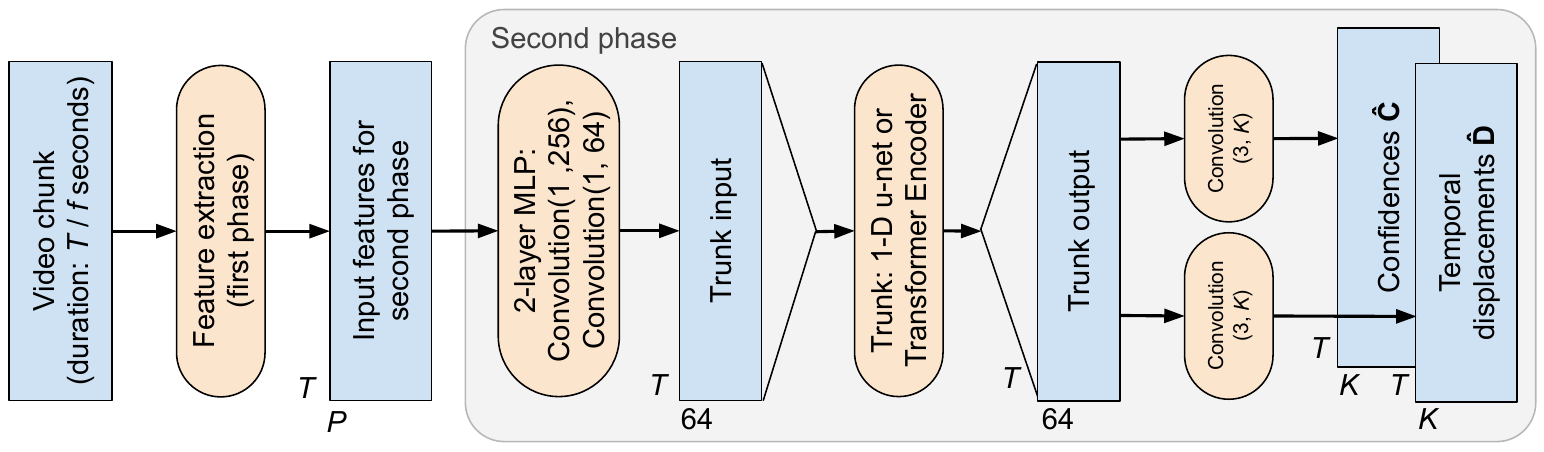}
  \caption{Architecture overview}
  \label{fig:diagram}
\end{subfigure}%
\hfill
\begin{subfigure}{0.3\linewidth}
  \centering
  \includegraphics[width=5.3cm]{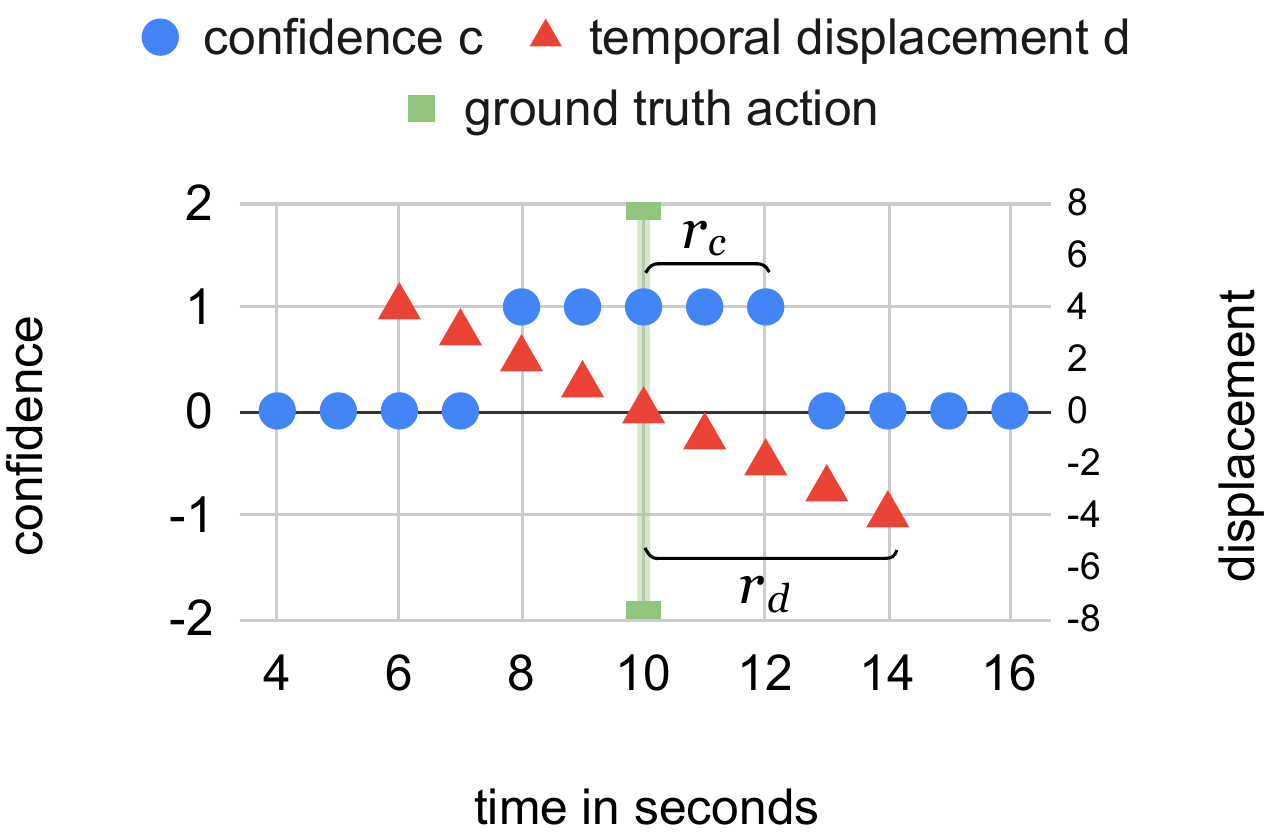}
  \caption{Targets around a ground-truth action}
  \label{fig:targets}
\end{subfigure}
\caption{\subref{fig:diagram} Architecture overview, where Convolution$(w,n)$ denotes a convolution with kernel size $w$ and $n$ filters. \subref{fig:targets}~Example of target confidences $c_{*,k}$ and temporal displacements $d_{*,k}$ for a single class $k$, around a ground-truth action of the same class, at one anchor per second. Note temporal displacement targets are undefined outside of the radius $r_d$ of the ground-truth action.}
\label{fig:methods}
\end{figure*}

Our solution, illustrated in Fig.~\ref{fig:methods}, makes use of a dense set of detection anchors. We define an anchor as a pair formed by a time instant (usually taken every 0.5 or 1.0 seconds) and action class, thus adopting a multi-label formulation. For each anchor, we predict both a detection confidence 
and a fine-grained temporal displacement. 
This approach leads to a new state-of-the-art on SoccerNet-v2. Experiments show large improvements in temporal precision, with a substantial benefit from the temporal displacements. In addition, this approach was later used as the basis for the first-place submission to the Action Spotting SoccerNet Challenge 2022~\cite{soares2022action}.

Our approach is inspired by work in object detection. Lin et al.~\cite{lin2017focal} demonstrated 
the advantages of using a dense set of detection anchors, with their single-stage RetinaNet detector surpassing the accuracy of slower contemporary two-stage counterparts. An important difference here is that the output space of action spotting is inherently lower-dimensional than that of object detection, as each action can be completely defined by its time and class. This allows us to use a very dense set of action spotting anchors at a relatively much lower cost.

For the trunk of our models, we experiment with a one-dimensional version of a u-net~\cite{ronneberger2015u} as well as a Transformer encoder~\cite{vaswani2017attention}. Both architectures incorporate large temporal contexts important for action spotting, while also preserving the smaller-scale features required for precise localization. We show that, while both architectures can achieve good results, the u-net has a better trade-off of time and accuracy.

SoccerNet-v2 
contains over 110K action 
labels.
We show that Sharpness-Aware Minimization (SAM)~\cite{foret2021sharpnessaware} and mixup data augmentation~\cite{zhang2018mixup} are able to improve results %
on the dataset, mitigating the lack of larger scale data or pretraining.


\section{Related Work}
\label{sec:related}

Since the release of the SoccerNet datasets~\cite{giancola2018soccernet,deliege2021soccernet}, 
several action spotting methods were proposed~\cite{tomei2021rms,cioppa2020context,cioppa2021camera,giancola2021temporally,zhou2021feature,vanderplaetse2020improved,vats2020event,chen2022faster}.

Concurrently with our work, Chen et al.~\cite{chen2022faster} developed a proposal-based (two-stage) temporal action detection model, which they then ported to the action spotting task.
While their approach has some similarities with ours, such as making use of temporal regression, our approach is simpler, consisting of a single stage designed specifically for action spotting.

Also related to our approach, RMS-Net~\cite{tomei2021rms} and CALF~\cite{cioppa2020context} adopt temporal regression, but with very different formulations.  RMS-Net~\cite{tomei2021rms} predicts at most one action per video chunk, and makes use of a max-pooling operation over the temporal dimension. It is thus not well suited for predicting multiple nearby actions. 
This differs from the CALF model~\cite{cioppa2020context}, which produces a set of possible action predictions per chunk, each of which 
may correspond to any time instant within the chunk, and belong to any class. The model is thus faced with a challenging problem to learn: simultaneously assigning all of its predictions to time instants and classes such that they cover all existing actions within the chunk. Our dense anchor approach sidesteps this challenge, by having each output anchor being preassigned to a time instant and class. This allows for predicting multiple actions per video chunk while using large chunk sizes that provide ample context for prediction. Our regressed temporal displacements are then used to finely localize each action in time. 

Zhou et al.~\cite{zhou2021feature} presented experiments with a Transformer Encoder (TE) on SoccerNet-v2. Differently from their work, we make use of the TE outputs at every position to produce dense anchor outputs. In addition, here we also experiment with a one-dimensional version of a u-net, showing that it has a better trade-off of time and accuracy relative to the TE.

\section{Methods}
\label{sec:methods}

Following previous works~\cite{deliege2021soccernet,tomei2021rms,cioppa2020context,giancola2021temporally,zhou2021feature}, we adopt a two-phase approach to action spotting, consisting of feature extraction followed by action prediction. This significantly decreases the computational burden during training and allows us to perform meaningful comparisons across methods. We note, however, that end-to-end training 
has shown to improve results~\cite{tomei2021rms} and represents a promising direction for future work.


Our two-phase architecture is illustrated in Fig.~\ref{fig:diagram}. In the first phase, a video chunk is decoded into frames, from which a sequence of $T$ feature vectors of dimension $P$ is extracted, composing a $T \times P$ feature matrix. In the second phase, this matrix is used to produce the action predictions. This starts with a single two-layer MLP applied independently to each input feature vector, resulting in a lower-dimensional output, which then gets fed into the model's {\it trunk}. As described in~\ref{sec:trunk}, the trunk combines information across all temporal locations while maintaining the sequence size $T$.
As described in~\ref{sec:anchors}, the trunk's output is used to create predictions for the dense set of $T \times K$ anchors, with $K$ the number of classes. When training, our loss is applied directly to all anchor predictions, while at test-time, post-processing is used to consolidate them into a set of action detections.


\subsection{Dense detection anchors}
\label{sec:anchors}

We use a dense set of detection anchors, inspired by dense single-stage object detectors~\cite{lin2017focal}. We define an anchor as a pair $(t, k)$, where $t = 1, 2, \dots, T$ indexes the $T$ temporal locations of a video chunk, and $k = 1, 2, \dots, K$ indexes the $K$ classes. The time difference between consecutive anchors is thus the same as that of consecutive input feature vectors, either 0.5 or 1.0 seconds in our experiments. To each anchor we associate a confidence and temporal displacement. We attached the output of our model's trunk to two heads, predicting respectively confidences $\hat{\mathbf{C}} = (\hat{c}_{t,k})$ and temporal displacements $\hat{\mathbf{D}} = (\hat{d}_{t,k})$. $\hat{\mathbf{C}}$ and $\hat{\mathbf{D}}$ are computed from the trunk outputs via their respective convolution operations, each with a temporal window of size 3 and $K$ output channels.

We define a confidence loss $L_c$ and a temporal displacement loss $L_d$, training a separate model for each rather than optimizing them jointly (see Section~\ref{sec:training}). The losses are computed with respect to {\it targets} (desired outputs), which are derived from the $N$ ground-truth actions contained within the given video chunk, which we denote $G=\{(t_i,k_i)\}_{i=1}^N$. These targets, illustrated in Fig.~\ref{fig:targets}, are described below.

The confidence loss $L_c$ for a video chunk 
is computed with respect to target confidences $\mathbf{C} = (c_{t,k})$, defined to be 1 within an $r_c$ seconds radius of a ground-truth action and 0 elsewhere, i.e. $c_{t,k} = I\left(\exists (s,k) \in G \colon \lvert s - t \rvert \leq r_c f\right)$, where $I$ is the indicator function and $f$ is the temporal feature rate (the number of feature vectors extracted per second). The confidence loss is defined as $L_c(\hat{\mathbf{C}}, \mathbf{C}) = \sum_{k=1}^K{\sum_{t = 1}^T{\text{CE}(\hat{c}_{t,k}, c_{t,k})}}$, where CE denotes the standard cross-entropy loss. We found that $r_c$ on the order of a few seconds gave the best results. This entails a loss in temporal precision, as the model learns to output high confidences within the whole radius of when an action actually happened, motivating the use of the temporal displacement outputs $\hat{\mathbf{D}}$. As we show in experiments, incorporating the displacements results in a large improvement to temporal precision.

The temporal displacement loss $L_d$ is only applied within an $r_d$ seconds radius of ground-truth actions, given that predicted displacements will only be relevant when paired with high confidences. Thus, for each class $k$, we first define its temporal support set $S(k) = \{t =1,2,\dots,T \mid \exists(s,k) \in G \colon \lvert s - t \rvert \leq r_d f\}$. We then define the loss $L_d(\hat{\mathbf{D}}, \mathbf{D}) = \sum_{k=1}^K \sum_{t \in S(k)}{L_h(\hat{d}_{t,k}, d_{t,k})}$, where $L_h$ denotes the Huber regression loss and the targets $\mathbf{D} = (d_{t,k})$ are defined so that each $d_{t,k}$ is the signed difference between $t$ and the temporal index of its nearest ground-truth action of class $k$ in $G$.

At test-time, to consolidate the predictions from $\hat{\mathbf{C}}$ and $\hat{\mathbf{D}}$, we apply two post-processing steps. The first displaces each confidence $\hat{c}_{t,k}$ by its corresponding displacement $\hat{d}_{t,k}$, keeping the maximum confidence when two or more are displaced into the same temporal location. The second step applies non-maximum suppression (NMS)~\cite{giancola2018soccernet,giancola2021temporally} to the displaced confidences. Since we adopt a multi-label formulation, we apply NMS separately for each class. To demonstrate the improvement from incorporating the temporal displacements, we later present an ablation where they are ignored, which is done simply by skipping the first post-processing step above. Note we do not apply any post-processing during training, instead defining the losses directly on the raw model predictions.

\subsection{Trunk architectures}
\label{sec:trunk}


We experiment with two trunk architectures.
The first is a 1-D u-net~\cite{ronneberger2015u}, which
consists of a contracting path that captures global context, followed by an expanding path, whose features are combined with those from the contracting path to enable precise localization. We replace the u-net's 2-D convolution blocks with \mbox{1-D} ResNet-V2 bottleneck blocks~\cite{he2016identity}, which improved results while stabilizing training.

The second trunk architecture we experiment with is a Transformer encoder (TE)~\cite{vaswani2017attention}, whose attention mechanism allows each token in a sequence to attend to all other tokens, thus incorporating global context while still preserving important local features. Relative to convolutional networks such as the u-net, Transformers have less inductive bias, often requiring pretraining on large datasets, or strong data augmentation and regularization~\cite{chen2021vision,devlin-etal-2019-bert}.
Here, we  achieve good results with the TE by training with Sharpness-Aware Minimization (SAM)~\cite{foret2021sharpnessaware} and mixup~\cite{zhang2018mixup}, as described in Section~\ref{sec:training}.

\subsection{Training}
\label{sec:training}

We train our models from scratch using the Adam optimizer with Sharpness-Aware Minimization (SAM)~\cite{foret2021sharpnessaware}, mixup data augmentation~\cite{zhang2018mixup}, and decoupled weight decay~\cite{loshchilov2017decoupled}. 
SAM seeks wide minima of the loss function, which has been shown to improve generalization~\cite{foret2021sharpnessaware}, especially for small datasets and models with low inductive bias~\cite{chen2021vision}.
We do not apply batch normalization when training the u-net, 
finding that its skip connections 
were sufficient to stabilize training.

We found it convenient to train confidence prediction separately from temporal displacement regression, resulting in a two-step approach. This provides similar results to joint training, while simplifying experimental design. We first train a model that produces only confidences, by optimizing the confidence loss $L_c$ and making use of mixup data augmentation~\cite{zhang2018mixup}. We then train a second model that produces only temporal displacements, by optimizing $L_d$, but {\it without} applying mixup. Due to the temporal displacement loss only being defined within small windows around ground-truth actions, we found it difficult to effectively apply mixup when using it.

\section{Experimental settings}
\label{sec:experiments}

\label{sec:settings}


We present results on two sets of features. The first
consists of ResNet-152 features extracted at $f = 2$ fps~\cite{deliege2021soccernet}.
We experiment with the PCA version,
here denoted ResNet+PCA. The second set comes from a series of models fine-tuned on SoccerNet-v2~\cite{zhou2021feature},
which we denote as Combination, whose features are extracted at $f = 1$ fps. Our two-layer MLP has layers with respectively 256 and 64 output channels, generating a $T \times 64$ matrix irrespective of the input feature set size.


We experimentally chose a chunk size of 112s
and radii 
$r_c = 3$s and $r_d = 6$s.  We use an NMS suppression window of 20s, following~\cite{giancola2021temporally,zhou2021feature}. Training and inference speeds were measured using wall times on a cloud instance with a V100 vGPU, 48 vCPUs at 2.30GHz, and 256GiB RAM.

At each contracting step of the u-net, we halve the temporal dimension and double the channels. Expansion follows a symmetric design. We contract and expand 5 times when using ResNet+PCA ($T=224$), and 4 times when using Combination features ($T=112$), so in both cases the smallest temporal dimension becomes $224 / 2^5 = 112 / 2^4 = 7$. For the TE, we experiment with two sizes: Small and Base. Small has 4 layers, embedding size 128, and 4 attention heads, while Base has 12 layers, embedding size 256 and 8 attention heads. 
We use batch size 64 for Base, and our default of 256 for Small.


For each model, we find the best set of hyper-parameters on the validation set. To decrease costs, we optimize each hyper-parameter in turn, in the following order: learning rate; SAM's $\rho$ (when applicable); weight decay; and mixup $\alpha$ (when applicable).
We use a batch size of 256
and train for 1,000 epochs, where each epoch consists of 8,192 uniformly sampled video chunks.
We apply a linear decay to the learning rate and weight decay, so that the final decayed values (at epoch 1,000) are 1/100th of the respective initial values. We train each model five times and report average results.

We report results on SoccerNet's average-mAP metric, which uses tolerances $\delta=5, 10, \dots, 60$s, as well as the recent {\it tight} average-mAP metric, which uses $\delta=1, 2, 3, 4, 5$s~\cite{spottingchallenge}. The tolerance $\delta$ defines the time difference allowed between a detection and a ground-truth action such that the detection may still be considered a true positive. Thus, smaller tolerances enforce higher temporal precision. Note $\delta$ is unrelated to the radii $r_c$ and $r_d$, the latter only used during training.

\section{Results}
\label{sec:results}

A set of ablations is presented in Table~\ref{tab:ablations}, where DU, DTES and DTEB stand for the dense anchor model using respectively the u-net, Small TE and Base TE. We see a large improvement when applying SAM with the ResNet+PCA features, but a very small one when applying it with the Combination features, which were already fine-tuned on the same dataset. Mixup gives small improvements across both feature types.
DTEB+SAM+mixup achieves average-mAP similar to that of DU+SAM+mixup, but with a much longer training time and lower throughput. 
Recent techniques have reduced the computational demands of Transformers~\cite{tay2020long}, while pretraining is well-known to improve their results~\cite{chen2021vision,devlin-etal-2019-bert}, though we have not currently explored those directions.

\begin{table}
\scriptsize
\begin{center}
\begin{tabular}{c c c c c c} 
 \Xhline{2\arrayrulewidth}
 Method & Features & \makecell{Avg.-\\mAP} & \makecell{Chunks/\\second} & \makecell{Epoch\\time} & Params \\
 \Xhline{2\arrayrulewidth}
 DU & RN+PCA & 63.8 & 1,447 & 5.4s & 17.5M \\ 

 DU+SAM & RN+PCA & 72.0 & 1,447 & 8.1s & 17.5M \\

 DU+SAM+mixup & RN+PCA & 72.2 & 1,447 & 8.3s & 17.5M \\
 \hline

 DTES+SAM+mixup & RN+PCA & 68.2 & 1,329 & 15.0s & 1.0M  \\
 \hline

 DTEB+SAM+mixup & RN+PCA & 72.4 & 342 & 112.0s & 9.7M  \\
 \hline
 DU & Combin. & 75.7 & 511 & 13.7s & 8.9M  \\
 
 DU+SAM & Combin. & 76.1 & 511 & 18.4s & 8.9M  \\
 
 DU+SAM+mixup & Combin. & 77.3 & 511 & 22.7s & 8.9M  \\
 \Xhline{2\arrayrulewidth}
\end{tabular}
\end{center}
\caption{\label{tab:ablations}Ablations exploring different feature sets, model trunks, and the use of SAM and mixup. 
Chunks/second measures each model's inference throughput, excluding initial feature extraction and all post-processing.}
\end{table}

\begin{figure}[htb]
\centering
\hspace*{-0.5cm}
\includegraphics[width=6.9cm]{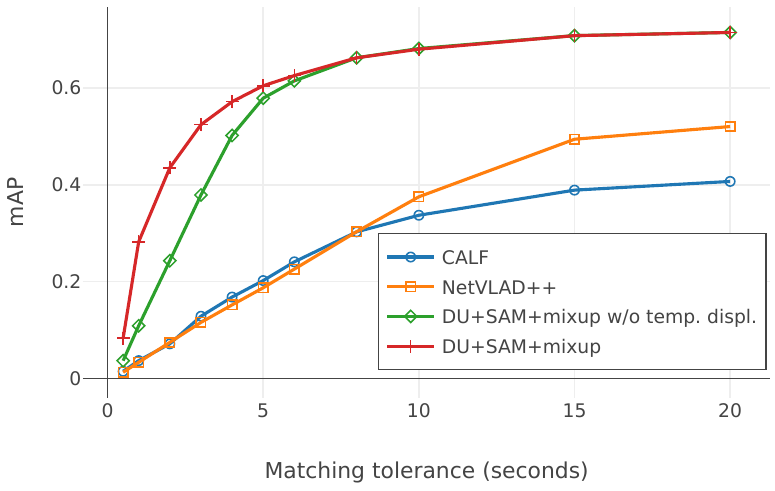}
\caption{mAP for different methods applied to ResNet+PCA features as a function of the matching tolerance $\delta$. While standard average-mAP is defined with tolerances of up to 60 seconds, the graph focuses on tolerances of up to 20 seconds.}
\label{fig:curves}
\end{figure}

Results comparing methods across various tolerances $\delta$ are presented in Figure~\ref{fig:curves}. We include results from CALF~\cite{cioppa2020context} and NetVLAD++~\cite{giancola2021temporally}, whose implementations were made available by their authors. 
All results in the figure were generated using the ResNet+PCA features. While our method outperforms the previous approaches across all tolerances, the improvement is significantly larger at smaller ones. The figure also shows that the temporal displacements provide significant improvements at small matching tolerances, without affecting results at larger ones. This observation is confirmed in Table~\ref{tab:comparisons}, where our method without the temporal displacements $\hat{\mathbf{D}}$ has much lower tight average-mAP.

Comparisons to prior work are presented in Table~\ref{tab:comparisons}. On the ResNet+PCA features, DU outperforms CALF~\cite{cioppa2020context} and NetVLAD++~\cite{giancola2021temporally}. Applying our model on Zhou et al.'s pre-computed features leads to a new state-of-the-art. Generally, our improvements are larger on the tight average-mAP metric. Our approach was further improved in later work~\cite{soares2022action}, leading to an improved state-of-the-art as well as the first-place in the Action Spotting SoccerNet Challenge 2022.

\begin{table}
\footnotesize
\begin{center}
\begin{tabular}{c c c c c} 
\Xhline{2\arrayrulewidth}
 Method & Features & A-mAP & Tight a-mAP \\
 \Xhline{2\arrayrulewidth}
 NetVLAD++~\cite{giancola2021temporally} & ResNet & 53.4 & 11.5$^*$ \\ 
 AImageLab RMSNet~\cite{tomei2021rms} & ResNet-tuned & 63.5$^*$ & 28.8$^*$ \\
 Vis. Analysis of Humans & CSN-tuned & 64.7$^\dag$ & 46.2$^\dag$ \\
 Faster-TAD~\cite{chen2022faster} & Swin-tuned & N/A & 54.1\\
 \hline
 CALF~\cite{deliege2021soccernet,cioppa2020context} & ResNet+PCA & 40.7 & 12.2$^\ddagger$ \\
 NetVLAD++~\cite{giancola2021temporally} & ResNet+PCA & 50.7 & 11.3$^\ddagger$\\
 DU & ResNet+PCA & 63.8 & 41.7 \\
 DU+SAM+mixup w/o $\hat{\mathbf{D}}$ & ResNet+PCA & 72.1 & 39.3\\
 DU+SAM+mixup & ResNet+PCA & {\bf 72.2} & {\bf 50.7}\\
 \hline
 Zhou et al.~\cite{zhou2021feature} & Combination & 73.8 & 47.1$^*$ \\
 DU & Combination & 75.7 & 58.5 \\
 DU+SAM+mixup w/o $\hat{\mathbf{D}}$ & Combination & {\bf 77.3} & 46.8\\
 DU+SAM+mixup & Combination & {\bf 77.3} & {\bf 60.7} \\
 \Xhline{2\arrayrulewidth}
\end{tabular}
\end{center}
\caption{\label{tab:comparisons}Comparison with results from prior works. *~Results reported on challenge website~\cite{spottingchallenge}. \dag~Results reported on challenge website for the {\it challenge} split, whereas all other reported results are on the standard {\it test} split. $\ddagger$ Results computed using the implementation provided by the authors.}
\end{table}

\section{Conclusion}
\label{sec:conclusion}

This work presented a temporally precise action spotting model that uses a dense set of detection anchors. The model sets a new state-of-the-art on SoccerNet-v2 with marked improvements when evaluated at smaller tolerances. For the model's trunk, we experimented with a 1-D u-net as well as a TE, showing that the TE requires a much larger computational budget to match the accuracy of the u-net.  Ablations demonstrated the importance of predicting fine-grained temporal displacements for temporal precision, as well as the benefits brought by training with SAM and mixup data augmentation.

\vspace{1mm}

\noindent {\small {\bf Acknowledgements} We are very grateful to Gaurav Srivastava for helpful discussions and for reviewing this work.}


\clearpage

\bibliographystyle{IEEEbib}
\bibliography{references}

\begin{thebibliography}{10}

\bibitem{giancola2018soccernet}
Silvio Giancola, Mohieddine Amine, Tarek Dghaily, and Bernard Ghanem,
\newblock ``{SoccerNet}: A scalable dataset for action spotting in soccer
  videos,''
\newblock in {\em Computer Vision and Pattern Recognition (CVPR) Workshops},
  2018.

\bibitem{deliege2021soccernet}
Adrien Deliege, Anthony Cioppa, Silvio Giancola, Meisam Seikavandi, Jacob
  Dueholm, Kamal Nasrollahi, Bernard Ghanem, Thomas Moeslund, and Marc
  Van~Droogenbroeck,
\newblock ``{SoccerNet-v2}: A dataset and benchmarks for holistic understanding
  of broadcast soccer videos,''
\newblock in {\em Computer Vision and Pattern Recognition (CVPR) Workshops},
  2021, pp. 4508--4519.

\bibitem{tomei2021rms}
Matteo Tomei, Lorenzo Baraldi, Simone Calderara, Simone Bronzin, and Rita
  Cucchiara,
\newblock ``{RMS-Net}: Regression and masking for soccer event spotting,''
\newblock in {\em International Conference on Pattern Recognition (ICPR)},
  2021.

\bibitem{cioppa2020context}
Anthony Cioppa, Adrien Deliege, Silvio Giancola, Bernard Ghanem, Marc~Van
  Droogenbroeck, Rikke Gade, and Thomas Moeslund,
\newblock ``A context-aware loss function for action spotting in soccer
  videos,''
\newblock in {\em Computer Vision and Pattern Recognition (CVPR)}, 2020.

\bibitem{cioppa2021camera}
Anthony Cioppa, Adrien Deliege, Floriane Magera, Silvio Giancola, Olivier
  Barnich, Bernard Ghanem, and Marc Van~Droogenbroeck,
\newblock ``Camera calibration and player localization in {SoccerNet-v2} and
  investigation of their representations for action spotting,''
\newblock in {\em Computer Vision and Pattern Recognition (CVPR) Workshops},
  2021, pp. 4537--4546.

\bibitem{giancola2021temporally}
Silvio Giancola and Bernard Ghanem,
\newblock ``Temporally-aware feature pooling for action spotting in soccer
  broadcasts,''
\newblock in {\em Computer Vision and Pattern Recognition (CVPR) Workshops},
  2021.

\bibitem{zhou2021feature}
Xin Zhou, Le~Kang, Zhiyu Cheng, Bo~He, and Jingyu Xin,
\newblock ``Feature combination meets attention: {Baidu} soccer embeddings and
  {Transformer} based temporal detection,''
\newblock {\em arXiv preprint arXiv:2106.14447}, 2021.

\bibitem{vanderplaetse2020improved}
Bastien Vanderplaetse and Stephane Dupont,
\newblock ``Improved soccer action spotting using both audio and video
  streams,''
\newblock in {\em Computer Vision and Pattern Recognition (CVPR) Workshops},
  2020, pp. 896--897.

\bibitem{vats2020event}
Kanav Vats, Mehrnaz Fani, Pascale Walters, David~A Clausi, and John Zelek,
\newblock ``Event detection in coarsely annotated sports videos via parallel
  multi-receptive field 1{D} convolutions,''
\newblock in {\em Computer Vision and Pattern Recognition (CVPR) Workshops},
  2020, pp. 882--883.

\bibitem{soares2022action}
Jo{\~a}o~V.~B. Soares and Avijit Shah,
\newblock ``Action spotting using dense detection anchors revisited: Submission
  to the {SoccerNet} {Challenge} 2022,''
\newblock {\em arXiv preprint arXiv:2206.07846}, 2022.

\bibitem{lin2017focal}
Tsung-Yi Lin, Priya Goyal, Ross Girshick, Kaiming He, and Piotr Doll{\'a}r,
\newblock ``Focal loss for dense object detection,''
\newblock in {\em International Conference on Computer Vision}, 2017.

\bibitem{ronneberger2015u}
Olaf Ronneberger, Philipp Fischer, and Thomas Brox,
\newblock ``U-net: Convolutional networks for biomedical image segmentation,''
\newblock in {\em International Conference on Medical image computing and
  computer-assisted intervention (MICCAI)}. Springer, 2015, pp. 234--241.

\bibitem{vaswani2017attention}
Ashish Vaswani, Noam Shazeer, Niki Parmar, Jakob Uszkoreit, Llion Jones, Aidan
  Gomez, {\L}ukasz Kaiser, and Illia Polosukhin,
\newblock ``Attention is all you need,''
\newblock in {\em Advances in neural information processing systems}, 2017.

\bibitem{foret2021sharpnessaware}
Pierre Foret, Ariel Kleiner, Hossein Mobahi, and Behnam Neyshabur,
\newblock ``Sharpness-aware minimization for efficiently improving
  generalization,''
\newblock in {\em International Conference on Learning Representations}, 2021.

\bibitem{zhang2018mixup}
Hongyi Zhang, Moustapha Cisse, Yann~N. Dauphin, and David Lopez-Paz,
\newblock ``mixup: Beyond empirical risk minimization,''
\newblock in {\em International Conference on Learning Representations (ICLR)},
  2018.

\bibitem{chen2022faster}
Shimin Chen, Chen Chen, Wei Li, Xunqiang Tao, and Yandong Guo,
\newblock ``Faster-{TAD}: Towards temporal action detection with proposal
  generation and classification in a unified network,''
\newblock {\em preprint arXiv:2204.02674}, 2022.

\bibitem{he2016identity}
Kaiming He, Xiangyu Zhang, Shaoqing Ren, and Jian Sun,
\newblock ``Identity mappings in deep residual networks,''
\newblock in {\em European conference on computer vision (ECCV)}, 2016.

\bibitem{chen2021vision}
Xiangning Chen, Cho-Jui Hsieh, and Boqing Gong,
\newblock ``When {Vision Transformers} outperform {ResNets} without
  pre-training or strong data augmentations,''
\newblock in {\em International Conference on Learning Representations (ICLR)},
  2022.

\bibitem{devlin-etal-2019-bert}
Jacob Devlin, Ming-Wei Chang, Kenton Lee, and Kristina Toutanova,
\newblock ``{BERT}: Pre-training of deep bidirectional transformers for
  language understanding,''
\newblock in {\em North American Association for Computational Linguistics
  (NAACL)}, 2019, pp. 4171--4186.

\bibitem{loshchilov2017decoupled}
Ilya Loshchilov and Frank Hutter,
\newblock ``Decoupled weight decay regularization,''
\newblock in {\em International Conference on Learning Representations (ICLR)},
  2019.

\bibitem{spottingchallenge}
``{SoccerNet} action spotting challenge description and results,''
  \url{https://github.com/SoccerNet/sn-spotting}, 2021,
\newblock Accessed: February, 2022.

\bibitem{tay2020long}
Yi~Tay, Mostafa Dehghani, Samira Abnar, Yikang Shen, Dara Bahri, Philip Pham,
  Jinfeng Rao, Liu Yang, Sebastian Ruder, and Donald Metzler,
\newblock ``Long range arena : A benchmark for efficient {Transformers},''
\newblock in {\em International Conference on Learning Representations (ICLR)},
  2021.

\end{thebibliography}

\end{document}